\documentclass{article}
\usepackage[authoryear]{natbib}
\usepackage{url} 

\usepackage{doi}

\DeclareUnicodeCharacter{03BB}{\ensuremath{\lambda}}
\usepackage{iclr2025_conference,times}
\usepackage{algorithm}
\usepackage{algpseudocode}
\usepackage{float}

\usepackage{amsmath,amsfonts,bm}

\def\eqref#1{equation~\ref{#1}}

\def\1{\bm{1}}

\DeclareMathAlphabet{\mathsfit}{\encodingdefault}{\sfdefault}{m}{sl}
\SetMathAlphabet{\mathsfit}{bold}{\encodingdefault}{\sfdefault}{bx}{n}

\usepackage{hyperref}
\usepackage{url}
\usepackage{placeins} 

\usepackage{graphicx}
\usepackage{subcaption} 
\usepackage{hyperref}
\usepackage{booktabs}
\usepackage{siunitx}
\usepackage{multirow}

\title{Can you Finetune your Binoculars? Embedding Text Watermarks into the Weights of Large Language Models}

\author{%
Fay Elhassan \\
Max Planck Institute for Intelligent Systems \\
ELLIS Institute Tübingen \\
\texttt{fay.elhassan@tue.ellis.eu} \\
\And
Niccolò Ajroldi \\
Max Planck Institute for Intelligent Systems \\
ELLIS Institute Tübingen \\
\texttt{niccolo.ajroldi@tue.ellis.eu} \\
\And
Antonio Orvieto \\
Max Planck Institute for Intelligent Systems \\
ELLIS Institute Tübingen \\
Tübingen AI Center \\
\texttt{antonio.orvieto@tue.ellis.eu} \\
\And
Jonas Geiping \\
Max Planck Institute for Intelligent Systems \\
ELLIS Institute Tübingen \\
Tübingen AI Center \\
\texttt{jonas@tue.ellis.eu} \\
}

\iclrfinalcopy 
\begin{document}

\maketitle

\begin{abstract}

\looseness -1 The indistinguishability of AI-generated content from human text raises challenges in transparency and accountability. While several methods exist to watermark models behind APIs, embedding watermark strategies directly into model weights that are later reflected in the outputs of the model is challenging. 
In this study we propose a strategy to finetune a pair of low-rank adapters of a model, one serving as the text-generating model, and the other as the detector, so that
%
%
 a subtle watermark is embedded into the text generated by the first model and simultaneously optimized for detectability by the second. In this way, the watermarking strategy is fully learned end-to-end. This process imposes an optimization challenge, as balancing watermark robustness, naturalness, and task performance requires trade-offs. We discuss strategies on how to optimize this min-max objective and present results showing the effect of this modification to instruction finetuning.



\end{abstract}

\section{Introduction}
Large language models (LLMs) have become a cornerstone of natural language processing tasks, achieving human-like fluency across a range of applications. However, the indistinguishability of AI-generated text from human-created content raises critical questions about transparency, ethical use, and security. As these models increasingly produce content that is widely disseminated, it becomes crucial to ensure accountability, particularly in scenarios involving misinformation, content moderation, or intellectual property disputes.

Embedding watermarks in AI-generated text has emerged as a potential solution to address these challenges. Watermarks enable traceability by embedding patterns into generated text that can later be identified, providing a mechanism for verifying the origin of the content. Originally, most watermark methods embed imperceptible, but identifiable patterns into existing text that may be generated from a language model. While effective, these post-hoc approaches introduced detectable artifacts, making them susceptible to adversarial removal or reducing linguistic naturalness. In contrast, contemporary watermarking techniques integrate watermark patterns during the text generation process itself, influencing token selection and model outputs without requiring post-processing. This approach enhances robustness and detectability while maintaining fluency, making it more suitable for high-stakes applications. However, because contemporary watermarking methods modify the sampling, they cannot be easily embedded into the weights of open-source models. The code can be provided with the open-source model, but users can simply choose not to apply the watermark. 

This paper introduces a novel methodology for embedding subtle watermarks directly during the training process into the weights of the model. We leverage the Binoculars score \citet{hans2024spotting} to measure detectability and train models during instruction tuning to maximize detectability according to this objective. To do so, we train a two-model approach that uses two LLaMA 3.1 models with Low-Rank Adaptation (LoRA). By embedding traceability directly into the model weights during fine-tuning, our method ensures that the generated output remains natural and that the watermark method is embedded directly into the weights of the model. This approach enables watermarking at the model level rather than post-processing or modifying the sampling process. While the final approach is still a proof-of-concept, we hope that this is a first step towards addressing challenges in ethical AI deployment and secure content generation for open-source models.

\section{Background and Related Work}
The field of watermarking AI-generated text has gained prominence in recent years, with substantial efforts focused on developing methods to ensure traceability and verifiability of AI outputs. This section reviews recent related work and contextualizes our contributions.

\subsection{Watermarking Methods}

Watermarking LLM-generated text has been extensively studied, with early approaches mostly on post-hoc watermarking of already existing text  \citet{atallah_natural_2001}. Current watermarking strategy instead watermark text as it is being generating, relying on specialized decoding and sampling strategies. One example is \citet{kirchenbauer2023watermarking}, who introduce a watermarking technique that modifies the token sampling algorithm by altering next-token probabilities in a pseudo-random manner based on a hash of local context.

While these generation-time watermarking techniques are effective, there is a natural trade-off between the magnitude of the distributional shift caused by the watermark, and its detectability \citet{kirchenbauer_reliability_2023}. Further, there has been a large-scale exploration of the susceptibility of watermarks to natural paraphrasing attacks and strong adversarial attacks \citet{krishna_paraphrasing_2023,sadasivan_can_2023,zhang2023watermarks}. A good amount of work has proposed new generation-time watermarks and modifications to existing schemes to improve in detection and generation performance \citet{fernandez2023three}, for example \citet{christ_undetectable_2023} and \citet{hu2023unbiased} who propose a unbiased watermarking techniques that ensures the watermarking process does not alter the probability distribution of generated text, in certain technical notions of imperceptible. 
Another approach, \citet{kuditipudi2023robust}, proposes a distortion-free watermarking approach that pre-samples a random key for LLM generation, with advantages especially on detection after paraphrasing. Cryptographic methods have also been explored, such as the private key-based undetectable watermarking approach in \citet{christ2023undetectable}, and rejection sampling methods for publicly verifiable watermarks \citet{fairoze2023publicly}.
However, inference-time watermarking methods still face key limitations, such as their inability to be publicly verifiable \citet{ajith2023performance} and their reliance on modifying 
sampling strategies at generation time, making them unsuitable for open-source deployment within the weights of a model.
\citet{abdelnabi2021adversarial} proposed the Adversarial Watermarking Transformer (AWT), which jointly trains an encoder-decoder with adversarial optimization to embed watermarks in text. Unlike statistical watermarking, AWT learns dynamic word substitutions, ensuring semantic coherence while resisting detection.

This key limitation of inference-time watermarking techniques on their dependence on controlled sampling strategies, arises because these strategies are primarily designed for API-based models where the model author has full control over token selection. With the rise of open-weight models, where users can freely modify sampling strategies, there need to be more watermarking methods that embed signals directly into the model weights rather than relying on post-hoc modification of model probabilities.
\citet{wong2024endtoend} introduce the first end-to-end logits-based watermarking method for LLMs, where both encoder and decoder networks are jointly optimized to enhance robustness and text quality their approach optimizes watermark embedding directly within the model’s training pipeline. By leveraging an online prompting technique to bypass non-differentiable operations.

To address these challenges, \citet{gu2023learnability} proposed watermark distillation, where LLMs learn to replicate decoding-based watermarking, allowing intrinsic watermark generation. Although effective, they found that fine-tuning on normal text weakens watermarking, and learning low-distortion watermarks demands significant data. This approach enables model-level watermarking, embedding watermarks directly into the model without inference-time modifications. More recently, \citet{xu2024learning} 
developed a Reinforcement Learning (RL)-based watermark,
where an RL agent optimizes watermark embedding by balancing detectability with text utility. In this framework, the LLM is trained as an RL agent to maximize a reward function that incorporates watermark detectability while maintaining text fluency. Despite its advantages, RL-based watermarking requires extensive computational resources and relies on complex reinforcement learning objectives, making it challenging to tune effectively.

Our framework builds upon the work of \citet{hans2024spotting}, who introduce the \textit{binoculars} objective, a metric that measures the likelihood that text is machine-generated by evaluating the statistics of the text using two related LLMs. In our work, we take this metric as a starting point and specifically train two LoRA adapters so that they fill the roles of the two models using in the binoculars objective. This way we can improve the detectability of machine-generated text based on this metric directly during instruction finetuning, trading off watermark detectability and text fluency.

There is an interesting connection between our proposed method and modern fine-tuning approaches for aligning LLM outputs with desired behaviors. Recent strategies, such as Direct Preference Optimization (DPO) \citet{rafailov2023direct} and Generalized Reinforcement Preference Optimization (GRPO) \citet{tang2024generalized}, have explored optimization techniques in this domain. DPO enhances language models by directly maximizing the likelihood of preferred responses while minimizing divergence from dispreferred ones, eliminating the need for reward modeling. GRPO extends this framework further by incorporating structured constraints, offering greater flexibility in fine-tuning objectives. Our method optimizes an implicit alignment objective, but instead of human preferences, we guide the model towards embedding detectable yet unobtrusive watermark signals.

\subsection{Contributions of This Work}

Most watermarking techniques rely on statistical patterns, token control, or cryptographic signatures to embed signals into text. However, any robust evaluation metric used for detection can itself be repurposed as a watermarking signal if incorporated into the training process.

We leverage the binocular score \citet{hans2024spotting} as both a detection signal and a watermarking objective by fine-tuning the model to maximize this score on human text while minimizing it on generated text. This process enables implicit watermarking without modifying text structure, instead embedding a detection signature directly into the model's learning dynamics."

Our method embeds watermarks directly into model weights by optimizing a performer model to generate text with low perplexity while maximizing the binocular score against an observer model. Using the Binoculars framework , we jointly train both models to balance watermark robustness and text fluency. A regularized barrier objective prevents over-optimization, ensuring subtle yet detectable watermarks.
Our approach integrates watermark embedding and detection within the training process. The observer model provides real-time feedback, enforcing watermark constraints without compromising linguistic quality. This formulation improves detectability, and allows for seamless watermark verification without additional post-processing. 


\section{Methodology}

Next, we will describe our methodology in detail.

\subsection{Detecting Machine-Generated Text with Binoculars}
To embed robust and detectable watermarks without degrading text fluency, we optimize the binocular score introduced in Hans et \citet{hans2024spotting}. Such detection score leverages the difference in perplexity between two language models to identify AI-generated text, that measures the discrepancy between how the \textit{performer} model (\(\mathcal{M}_P\)) and \textit{observer} model (\(\mathcal{M}_O\)) assess the predictability of a generated sequence.
Given a sequence of text \(s = (x_1, \dots, x_L)\), we define the binocular score  as:
\[
    \mathcal{B}(s) = \frac{\log \operatorname{PPL}(s)}{\log \operatorname{XPPL}(s)},
\]
where:
\begin{align}
    \log \operatorname{PPL}(s) &= -\frac{1}{L} \sum_{i=1}^{L} \log \mathcal{M}_O(s[:i])_{x_i}, \\
    \log \operatorname{XPPL}(s) &= -\frac{1}{L} \sum_{i=1}^{L} \langle \mathcal{M}_O(s[:i]), \log \mathcal{M}_P(s[:i]) \rangle.
\end{align}

Here, \(\mathcal{M}(s[:i]) \in \mathbb{R}^{|V|}\) represents the probability distribution (softmax of logits) over the vocabulary \(V\) when predicting the \(i\)-th token of \(s\), conditioned on the text up to the $(i-1)$-th token: \(s[:i] = (x_1, ..., x_{i-1})\).
The log-perplexity (\(\operatorname{PPL}\)) quantifies how well the observer predicts the correct tokens, while the cross-perplexity (\(\operatorname{XPPL}\)) measures the alignment between the probability distributions of the observer and performer. A high binocular score indicates that the observer finds the text unexpected but well-aligned with the performer's outputs, signaling the presence of a watermark.

\subsection{Optimizing the Binoculars}

We optimize the binocular score by training both the performer (\(\mathcal{M}_P\)) and the observer model. We train the first to generate text and the second to detect generations from the first. Both models are initialized from a shared LLaMA 3.1 8B base and fine-tuned using Low Rank Adaptation (LoRA), which ensures efficient training while embedding watermarking patterns directly into the model weights.

A task-specific objective, modeled using standard cross-entropy loss, encourages fluent text generation:
\[
L_{\text{task}} = - \sum_{i=1}^{L} \log \mathcal{M}_P(x_i \mid x_{<i}),
\]

The second component of the loss is the binocular score \(L_{\text{binocular}} = \mathcal{B}(s)\), which encourages the performer model to generate text that is both natural and watermarked.

To balance both objectives, the total loss function is defined as:
\[
L_{\text{total}} = L_{\text{task}} + \lambda \cdot L_{\text{binocular}},
\]
where \( \lambda \) is a scaling factor that controls the contribution of the binocular loss relative to the task-specific loss.

\begin{algorithm}[h]
\caption{Optimization of Watermarked LLM}
\label{alg:watermark_training}
\begin{algorithmic}[1]
\Require Pre-trained Model, instruction data, optimizer, LoRA configurations
\State \textbf{Initialize:} Load pre-trained model weights into performer model $\mathcal{M}_P$ and observer model $\mathcal{M}_O$
\State \textbf{Apply LoRA:} 
    \Statex \quad $\mathcal{M}_P$: $r=32$, $\alpha=128$, target layers: \texttt{q\_proj}, \texttt{k\_proj}, \texttt{v\_proj}, \texttt{o\_proj}
    \Statex \quad $\mathcal{M}_O$: $r=16$, $\alpha=32$, same target layers
\State \textbf{Set optimizer:} Initialize optimizer with learning rate $\eta$ and batch size $B$
\For{each training step}
    \State \textbf{Sample} batch $\mathcal{D} = \{s_{\text{real},1}, \dots, s_{\text{real},n}\}$ from dataset ultrachat\_200k
    \For{each sequence $s_{\text{real}} \in \mathcal{D}$}
        \State \textbf{Generate watermarked text:} $s_{\text{gen}} \sim \mathcal{M}_P(s_{\text{real}})$
        \State \textbf{Compute task loss:}
        \begin{equation}
        L_{\text{task}} = - \sum_{i=1}^{L} \log \mathcal{M}_P(x_i \mid x_{<i}, s_{\text{real}})
        \end{equation}
        \State \textbf{Compute binocular scores on both $s_\text{real}$ and $s_\text{gen}$:}
        \begin{equation}
        \mathcal{B}(s) = 
        \frac{-\frac{1}{L} \sum_{i=1}^{L} \langle 1_{x_i}, \log(\mathcal{M}_P(s)[:i]))\rangle}
        {-\frac{1}{L} \sum_{i=1}^{L} \langle \mathcal{M}_O(s[:i]), \log (\mathcal{M}_P(s[:i])) \rangle}
        \end{equation}
        \State \textbf{Compute total loss:}
        \begin{equation}
        L_{\text{total}} = L_{\text{task}} - \lambda \left( \mathcal{B}(s_{\text{real}})
        + \mathcal{B}(s_{\text{gen}})) \right)
        \end{equation}
        \State \textbf{Backpropagate and update parameters} using optimizer
    \EndFor
\EndFor
\State \textbf{Return} trained models $\mathcal{M}_P$ and $\mathcal{M}_O$
\end{algorithmic}
\end{algorithm}


With this training objective, we can set up a min-max optimization framework that integrates the observer model’s assessment of watermark detectability into the learning process. We maximize the binocular score on the text generated by the performer model while minimizing it on real human text. Formally, we define:

\begin{itemize}
    \item \(s_{\text{real}}\): A real human-written sequence sampled from the dataset ultrachat\_200k.
    \item \(s_{\text{gen}}\): A watermarked sequence generated by the performer model \(\mathcal{M}_P\).
\end{itemize}

This leads to the following optimization objective:

\begin{equation}\label{eq:min-max}
    L_{\mathcal{M}_P}(s_{\text{real}}) - \lambda \left( \mathcal{B}_{\mathcal{M}_O, \mathcal{M}_P}(s_{\text{real}}) + \mathcal{B}_{\mathcal{M}_O, \mathcal{M}_P}(s_{\text{gen}}) \right)
\end{equation}

where the task loss \(L_{\mathcal{M}_P}(s_{\text{real}})\) ensures fluency and coherence, while the binocular scores \(\mathcal{B}_{\mathcal{M}_O, \mathcal{M}_P}\) regulate watermark detectability by distinguishing human text from watermarked outputs.







\subsection{From Min-Max Optimization to Regularized Training}
While the objective in Eq.~\ref{eq:min-max} is trainable, it suffers from the fact that the binoculars score is defined as a ratio, making it hard to balance the unlimited upside of maximizing the score value with the limited upside of reducing the objective further. To mitigate this, we reformulate into a constrained optimization problem.

The adversarial nature of the interaction between the performer and observer models can also lead to oscillatory training behavior, where the model alternates between embedding overly strong and overly weak watermarks. Additionally, hard constraints in min-max training make optimization difficult, often resulting in poor convergence and an imbalanced trade-off between watermark detectability and text quality. Further alternative formulations and experimental results exploring different constraints and optimization techniques are detailed in the appendix.\ref{appendix:optimization}




\subsection{Regularized Optimization for Watermark Embedding}

To balance text naturalness and watermark robustness in a more stable way, we reformulate into a constrained optimization problem, stabilizing training by controlling the performer's loss function, preventing excessive optimization that could degrade text fluency or result in overly detectable watermarks.

\begin{equation}
    \min_{\mathcal{M}_P, \mathcal{M}_O} \mathcal{B}_{\mathcal{M}_O, \mathcal{M}_P}(s_\text{real}) - \mathcal{B}_{\mathcal{M}_O, \mathcal{M}_P}(s_\text{gen})  \quad \text{subject to} \quad L_{\mathcal{M}_P}(s_\text{real}) \leq L_0.
\end{equation}
Here, the cross-entropy loss is constrained to remain below \( L_0 \), maintaining natural text generation quality. This formulation ensures that watermark detectability is maximized without compromising fluency.

\subsubsection{Solving the Constrained Optimization Problem}
To enforce the constraint, we utilize standard optimization theory, setting up \textbf{barrier functions}, which discourage violations by introducing penalties into the loss function. \citet{boyd2004convex}

The \textbf{exponential barrier function} applies a strong penalty as the constraint approaches violation:

\begin{equation}
    L_{\text{barrier}}^{\text{exp}} = \exp(L_{\mathcal{M}_P}(s_\text{real}) - L_0),
\end{equation}

where \( L_0 \) is a threshold that regulates updates. When \( L_{\mathcal{M}_P} \) exceeds \( L_0 \), optimization is strongly dampened to prevent destabilizing updates. 

Alternatively, the \textbf{quadratic barrier function} penalizes violations smoothly:

\begin{equation}
    L_{\text{barrier}}^{\text{quad}} = \max(L_{\mathcal{M}_P}(s_\text{real}) - L_0, 0)^2.
\end{equation}

This ensures that small deviations from \( L_0 \) receive minimal penalty, while larger deviations are penalized quadratically. Notably, neither formulation is a full interior-point method, as we want to allow some constraint violation, if necessary. The quadratic approach fully an exterior-point method, with the constraint being inactive while it is not violated.

\section{Results and Evaluation}
\subsection{Experimental Design and Optimization}
We evaluate our proposed approach to train models while embedding watermarks empirically. To do so, we fine-tune a LLaMA 3.1 8B model on the \texttt{HuggingFaceH4/ultrachat\_200k} dataset, which provides a diverse set of instruction-response pairs for alignment training. 
\paragraph{Optimization and LoRA Configuration}  
We train using the AdamW optimizer with weight decay, a learning rate \(\eta\), and batch size \(B\) with gradient accumulation. LoRA is applied to reduce training overhead, keeping most weights frozen. The observer model uses a larger LoRA (\(r=32, \alpha=128\)) for better watermark detection, while the performer model employs a smaller LoRA (\(r=16, \alpha=32\)) to preserve fluency. LoRA is applied to attention layers (\texttt{q\_proj}, \texttt{k\_proj}, \texttt{v\_proj}, \texttt{o\_proj}) to optimize efficiency.

\paragraph{Experimental Setup}
The evaluation spans diverse datasets to assess watermark embedding effectiveness. For general knowledge and news, we use CC-News, CNN/DailyMail, and Wikipedia. Commonsense reasoning is tested with OpenBookQA and AI2-ARC. Additionally, Databricks-Dolly-15K evaluates response quality in instruction-tuned models, ensuring broad coverage of text generation tasks.



\subsection{Binocular Score Analysis}
\begin{figure}[t]
    \centering
    \begin{minipage}{0.48\textwidth}
        \centering
        \includegraphics[width=\textwidth]{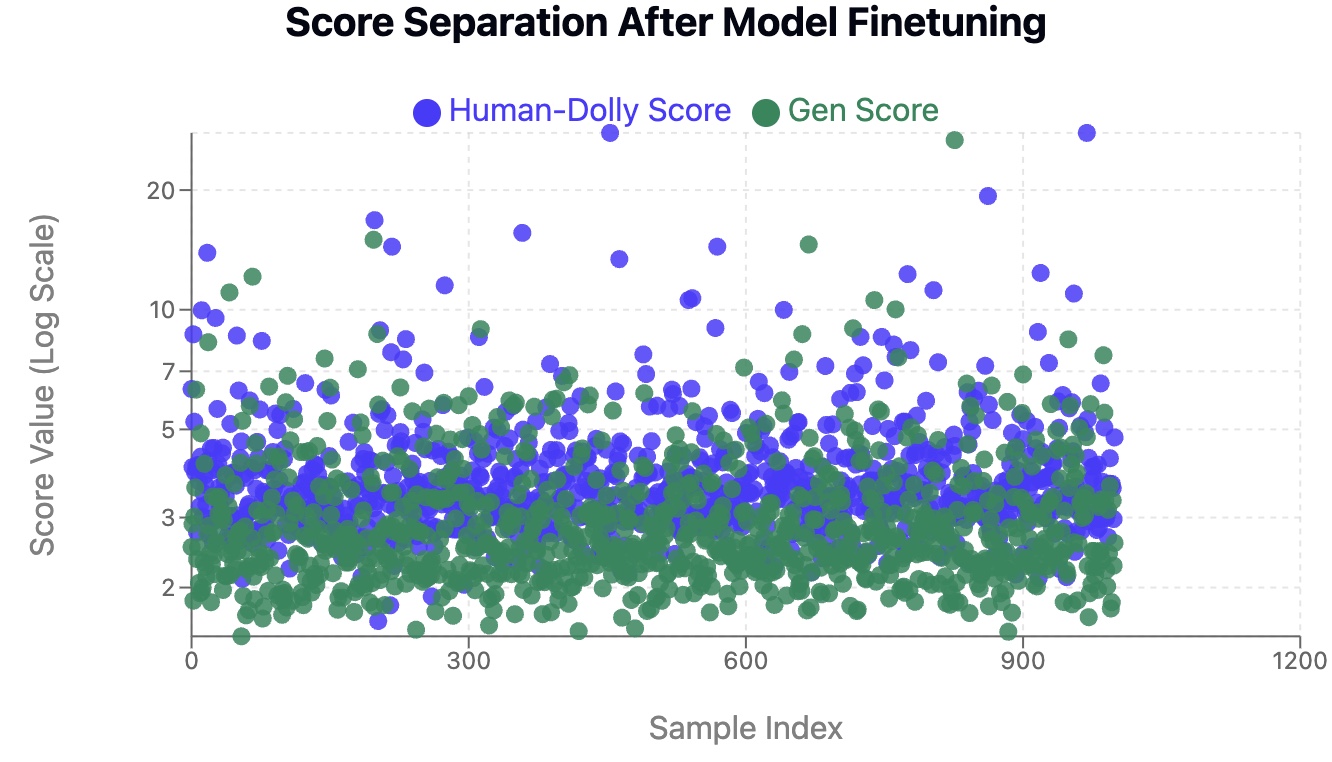}
        \subcaption{Training without regularizer (\(\lambda = 1e^{-2}\))}
    \end{minipage}
    \hspace{0.02\textwidth} 
    \begin{minipage}{0.45\textwidth}
        \centering
        \includegraphics[width=\textwidth]{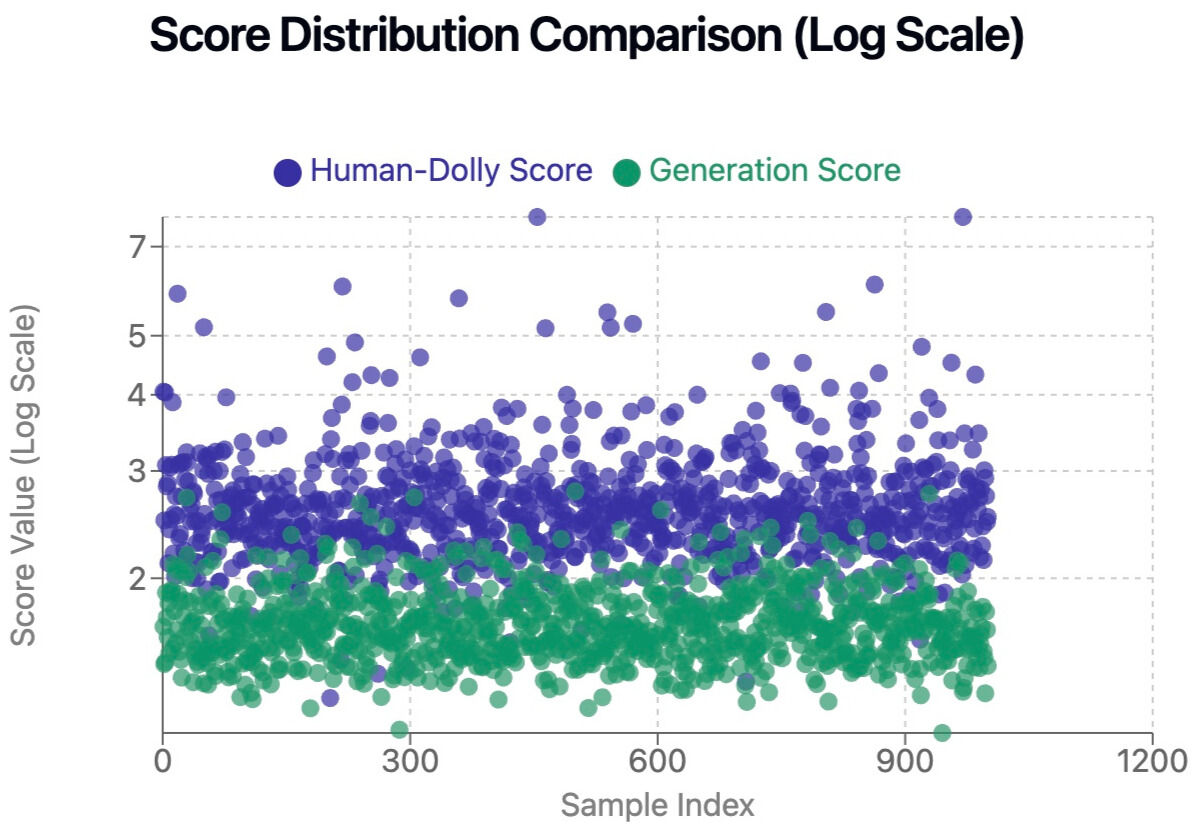}
        \subcaption{Exponential regularizer (\(\lambda = 1e^{-2}\))}
    \end{minipage}
    \caption{Scatter plots (log-scale) of binocular scores for human and generated text. (a) shows performance after fine-tuning without regularization, while (b) illustrates improved separation with the exponential regularizer.}
    \label{fig:binocular_scatter_combined}
    \vspace{-0.3cm}
\end{figure}

To evaluate the impact of fine-tuning on the separation between human and generated text, we analyze the distribution of binocular scores before and after applying the exponential regularizer with \(\lambda = 1e^{-2}\) (Figure \ref{fig:binocular_scatter_combined}). This analysis demonstrates how our watermark-embedded model differentiates generated content from human-written text while maintaining linguistic fluency.

The results highlight that the application of the exponential/quadratic regularizer enhances the distinguishability of watermarked text from human-written text. This suggests that fine-tuning with controlled regularization maintains the subtlety of the watermark while ensuring robustness against potential adversarial modifications.

\section{Evaluation and Results}

\subsection{Watermark Detection Performance}

To assess the effectiveness of our watermarking approach, we report detection performance using the Receiver Operating Characteristic (ROC) curves and Precision-Recall (PR) curves before and after fine-tuning. These metrics provide insight into the trade-offs between true positive rate (TPR), false positive rate (FPR), and precision-recall balance, highlighting how well the observer model detects watermarked text.

\begin{figure}[h]
    \centering
    \begin{minipage}{0.4\linewidth}
        \centering
        \includegraphics[width=\linewidth]{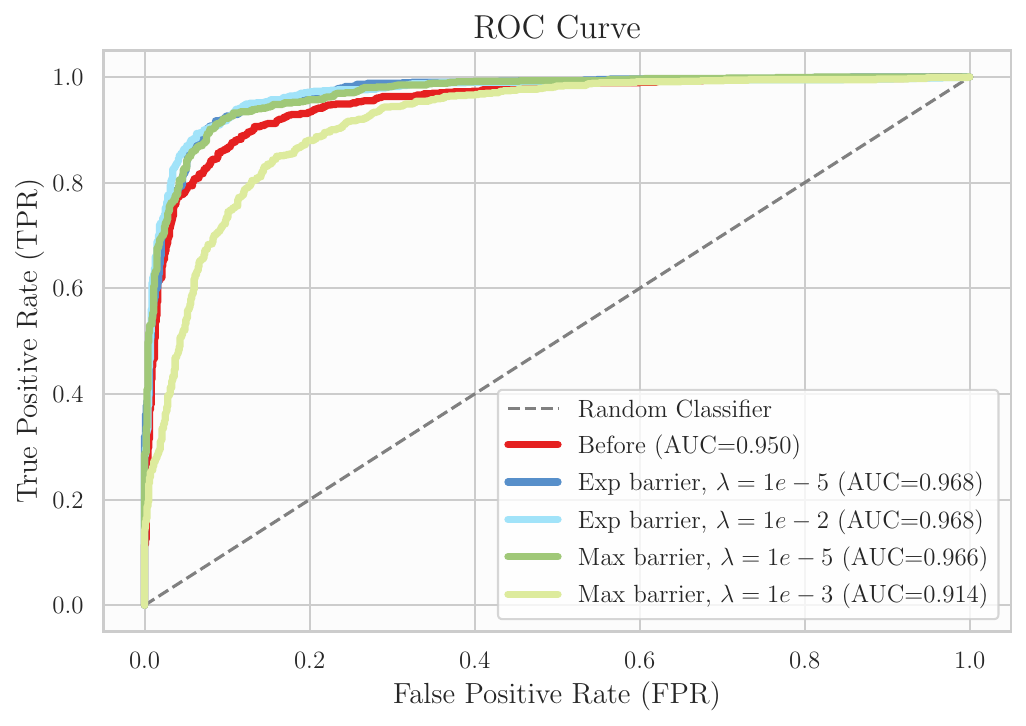}
        \subcaption{ROC curve comparison across different conditions. 
        }
        \label{fig:roc_curve}
    \end{minipage}
    \hspace{0.01\linewidth}
    \begin{minipage}{0.4\linewidth}
        \centering
        \includegraphics[width=\linewidth]{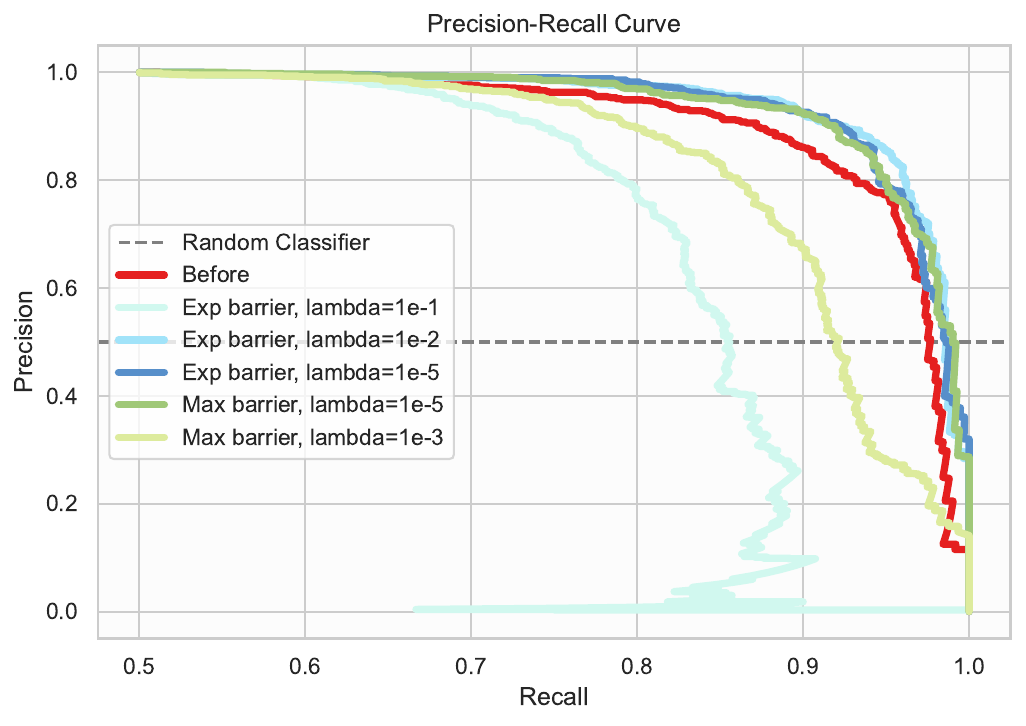}
        \subcaption{Precision-Recall curves comparing different experimental conditions. 
        }
        \label{fig:pr_curve}
    \end{minipage}
    \caption{Comparison of ROC and Precision-Recall curves for watermarking strategies. Exponential barriers improve detectability and maintain precision, while quadratic barriers degrade at higher recall thresholds.}
    \vspace{-.3cm}
    
\end{figure}

\subsection{Impact of Regularized Optimization on Detection Performance}

Figure~\ref{fig:roc_curve} highlights how models fine-tuned with regularized optimization barriers achieve higher TPR while maintaining lower FPR. Specifically, models trained with an exponential barrier function at \(\lambda = 1e^{-2}\) and \(\lambda = 1e^{-5}\) outperform the baseline, achieving an AUC-ROC of approximately 0.968, compared to 0.950 for the base model. This improvement confirms that fine-tuning effectively enhances watermark detectability without introducing excessive false positives.

PR curve analysis (Figure~\ref{fig:pr_curve}) further supports this finding. Unlike ROC curves, which assess classification performance across all decision thresholds, PR curves emphasize performance in scenarios with class imbalance, making them particularly relevant for watermark detection. Fine-tuned models exhibit higher precision across a broad recall range, confirming their ability to confidently identify watermarked text while minimizing false positives.

However, aggressively fine-tuned models (\(\lambda = 0.1\), labeled Exp, $\lambda=0.1$ and Small Lora Exp, $\lambda=0.1$) show a drop in precision at higher recall values. This suggests that while stronger watermarking enhances detectability, excessive optimization degrades fluency and increases false positives. This underscores the role of regularized barriers in maintaining the balance between watermark robustness and natural text generation.

\subsection{Effect of Regularized Barriers and Scaling Factors}

The influence of different regularized optimization barriers is summarized in Table~\ref{tab:ablation_study}, where we evaluate how \(\lambda\) scaling factors impact classification accuracy, detectability, and text fluency.

\FloatBarrier 

\begin{table}[h!]
    \centering
    \small
    \caption{Effect of Regularized Barriers on Performance Metrics}
    \label{tab:ablation_study}
    \renewcommand{\arraystretch}{0.7}
    \setlength{\tabcolsep}{2pt}
    \begin{tabular}{l c c c c}
        \toprule
        \textbf{Condition} & \textbf{Acc.} & \textbf{F1} & \textbf{ROC-AUC} & \textbf{PR-AUC} \\
        \midrule
        Baseline (Before FT)   & 0.886 & 0.888 & 0.950 & 0.950 \\
        Exp, $\lambda=1e\text{-}2$   & 0.915 & 0.913 & \textbf{0.968} & \textbf{0.969} \\
        Exp, $\lambda=1e\text{-}5$   & 0.915 & 0.915 & 0.968 & 0.968 \\
        Max, $\lambda=1e\text{-}2$   & 0.890 & 0.890 & 0.954 & 0.955 \\
        Max, $\lambda=1e\text{-}5$   & 0.912 & 0.913 & 0.966 & 0.967 \\
        Max, $\lambda=1e\text{-}3$   & 0.844 & 0.845 & 0.914 & 0.906 \\
        Exp, $\lambda=0.1$           & 0.801 & 0.813 & 0.868 & 0.829 \\
        Exp, $\lambda=1$             & 0.764 & 0.781 & 0.832 & 0.812 \\
        Max, $\lambda=1$             & 0.780 & 0.790 & 0.870 & 0.860 \\
        Exp, $\lambda=0.5$           & 0.799 & 0.805 & 0.861 & 0.832 \\
        Max, $\lambda=0.5$           & 0.810 & 0.820 & 0.900 & 0.892 \\
        \bottomrule
    \end{tabular}
\end{table}

\FloatBarrier 

Fine-tuned models with lower values of \(\lambda\) demonstrate improved watermark detectability without significantly degrading fluency. For instance:
- Exp $\lambda=1e\text{-}5$ and Max $\lambda=1e\text{-}5$ achieve classification accuracy of 0.9150 and 0.9120, respectively, with strong ROC-AUC scores (0.9679 and 0.9664).
- In contrast, models with higher \(\lambda\) values (Exp, $\lambda=0.1$ and Small \_ Exp, $\lambda=0.1$) experience a drop in accuracy (0.8005 and 0.7645, respectively), suggesting that excessive watermarking interferes with text fluency.

\subsection{Impact of LoRA Configuration on Watermarking Performance}

The size of the LoRA configuration affects both the performer and observer models, influencing watermark embedding and text fluency. Larger LoRA configurations enable stronger watermark separation while preserving text quality, whereas smaller configurations weaken watermarking, reducing classification accuracy.

\FloatBarrier 

\begin{table}[h!]
    \centering
    \small
    \caption{Impact of LoRA Configurations on Watermark Performance}
    \label{tab:lora_size_study}
    \renewcommand{\arraystretch}{0.8}
    \setlength{\tabcolsep}{2pt}
    \begin{tabular}{l c c c c}
        \toprule
        \textbf{Condition} & \textbf{LoRA Config} & \textbf{Acc.} & \textbf{F1} & \textbf{ROC-AUC} \\
        \midrule
        \textbf{Before FT} & \(r=32, \alpha=128\) & 0.886 & 0.888 & 0.950 \\
        \midrule
        Exp, $\lambda=1e\text{-}2$ (Full)   & \(r=32, \alpha=128\)  & 0.915 & 0.913 & \textbf{0.968} \\
        Exp, $\lambda=1e\text{-}5$ (Full)   & \(r=32, \alpha=128\)  & 0.915 & 0.915 & 0.968 \\
        Exp, $\lambda=1$ (Full)      & \(r=32, \alpha=128\)  & 0.764 & 0.780 & 0.832 \\
        Exp, $\lambda=0.5$ (Full)    & \(r=32, \alpha=128\)  & 0.799 & 0.805 & 0.861 \\
        Max, $\lambda=1e\text{-}2$ (Full)   & \(r=32, \alpha=128\)  & 0.890 & 0.890 & 0.954 \\
        Max, $\lambda=1$ (Full)      & \(r=32, \alpha=128\)  & 0.780 & 0.790 & 0.870 \\
        Max, $\lambda=0.5$ (Full)    & \(r=32, \alpha=128\)  & 0.810 & 0.820 & 0.900 \\
        \midrule
        Exp, $\lambda=1e\text{-}2$ (Small)  & \(r=16, \alpha=32\)   & 0.892 & 0.890 & 0.955 \\
        Exp, $\lambda=1e\text{-}5$ (Small)  & \(r=16, \alpha=32\)   & 0.900 & 0.901 & 0.960 \\
        Exp, $\lambda=1$ (Small)     & \(r=16, \alpha=32\)   & 0.750 & 0.765 & 0.825 \\
        Exp, $\lambda=0.5$ (Small)   & \(r=16, \alpha=32\)   & 0.780 & 0.790 & 0.845 \\
        Max, $\lambda=1e\text{-}2$ (Small)  & \(r=16, \alpha=32\)   & 0.870 & 0.875 & 0.930 \\
        Max, $\lambda=1e\text{-}5$ (Small)  & \(r=16, \alpha=32\)   & 0.880 & 0.882 & 0.940 \\
        Max, $\lambda=1$ (Small)     & \(r=16, \alpha=32\)   & 0.770 & 0.780 & 0.860 \\
        Max, $\lambda=0.5$ (Small)   & \(r=16, \alpha=32\)   & 0.800 & 0.810 & 0.875 \\
        \bottomrule
    \end{tabular}
    \vspace{-0.2cm}
\end{table}

\FloatBarrier 

This is evident in Table~\ref{tab:lora_size_study}, where models with smaller LoRA configurations show lower post-training accuracy and F1 scores, particularly in the exponential case with \(\lambda = 0.1\). Larger LoRA setups consistently improve watermark robustness without degrading fluency.

\subsection{Discussion on Model Performance with Watermark Fine-tuning}

The results in Table~\ref{tab:evaluation} demonstrate that fine-tuning with an exponential barrier watermark (\(\lambda = 1e^{-2}\)) leads to notable improvements in reasoning-heavy benchmarks while maintaining competitive accuracy across other tasks. Specifically, the watermarked model achieves a substantial increase in GSM8K (+8.0\%) and MMLU overall (+4.7\%), reinforcing that watermarking enhances structured reasoning and logical inference.Similarly, TruthfulQA (Gen) gains +5.7 BLEU, indicating that watermarking affects text constraints without harming factual consistency.
\\\
\begin{table*}[h]
\centering
\renewcommand{\arraystretch}{1.2}
\resizebox{\textwidth}{!}{%
\begin{tabular}{lcccccccc}
\toprule
\multirow{2}{*}{\textbf{Task}} & \multirow{2}{*}{\textbf{Metric}} & \multicolumn{2}{c}{\textbf{Baseline Model}} & \multicolumn{2}{c}{\textbf{Fine-tuned Non-Watermarked}} & \multicolumn{2}{c}{\textbf{Watermarked Model (λ=1e-2)}} \\
& & \textbf{Value} & \textbf{Stderr} & \textbf{Value} & \textbf{Stderr} & \textbf{Value} & \textbf{Stderr} \\
\midrule
\textbf{ARC-Challenge} & Accuracy & 0.5119 & ±0.0146 & 0.5128 & ±0.0146 & 0.5154 & ±0.0146 \\
\textbf{ARC-Easy} & Accuracy & 0.8106 & ±0.0080 & 0.8077 & ±0.0081 & 0.8169 & ±0.0079 \\
\textbf{GSM8K} & Exact Match & 0.6763 & ±0.0129 & 0.6816 & ±0.0128 & 0.7566 & ±0.0118 \\
\textbf{HellaSwag} & Accuracy & 0.5880 & ±0.0049 & 0.5884 & ±0.0049 & 0.5910 & ±0.0049 \\
\textbf{LAMBADA (OpenAI)} & Accuracy & 0.7452 & ±0.0061 & 0.7419 & ±0.0061 & 0.7312 & ±0.0062 \\
& Perplexity & 3.1566 & ±0.0636 & 3.1352 & ±0.0634 & 3.4046 & ±0.0731 \\
\textbf{MMLU} & Accuracy & 0.6498 & ±0.0038 & 0.6439 & ±0.0038 & 0.6807 & ±0.0037 \\
\textbf{MMLU-Humanities} & Accuracy & 0.5749 & ±0.0067 & 0.5681 & ±0.0066 & 0.6431 & ±0.0067 \\
\textbf{MMLU-STEM} & Accuracy & 0.5890 & ±0.0085 & 0.5772 & ±0.0085 & 0.5864 & ±0.0084 \\
\textbf{TruthfulQA (Gen)} & BLEU Score & 30.4135 & ±0.8623 & 28.6473 & ±0.8422 & 36.0994 & ±0.8775 \\
\textbf{TruthfulQA (MC1)} & Accuracy & 0.3378 & ±0.0166 & 0.3415 & ±0.0166 & 0.3721 & ±0.0169 \\
\textbf{TruthfulQA (MC2)} & Accuracy & 0.5026 & ±0.0147 & 0.4947 & ±0.0149 & 0.5408 & ±0.0149 \\
\textbf{PIQA} & Accuracy & 0.8003 & ±0.0093 & 0.8030 & ±0.0093 & 0.8041 & ±0.0092 \\
\bottomrule
\end{tabular}%
}
\caption{Evaluation results comparing the \textbf{Baseline Model}, \textbf{Fine-tuned Non-Watermarked Model}, and \textbf{Fine-tuned Watermarked Model (λ=1e-2)}. The watermarked model shows better performance on reasoning-heavy tasks (MMLU, GSM8K) while maintaining competitive accuracy elsewhere.}
\label{tab:evaluation}
\end{table*}

Across general commonsense reasoning tasks, including ARC-Challenge (+0.7\%), HellaSwag (+0.5\%), and PIQA (+0.4\%), performance remains stable, indicating that the watermarking mechanism does not significantly alter real-world knowledge application. However, a minor increase in perplexity (LAMBADA) suggests a slight trade-off in fluency due to watermarking constraints. Despite this, the overall results confirm that embedding watermarks directly into model weights does not degrade generalization. Instead, the fine-tuned watermarked model balances detectability and performance, proving that watermarking can serve as a dual-purpose mechanism for both model auditing and enhanced structured learning.

    
\subsection{Final Insights on Optimization Trade-offs}

Our findings confirm that fine-tuning with regularized barriers enables watermarking to be both effective and resilient, offering a scalable approach for watermarking AI-generated text without sacrificing usability. Models trained with \(\lambda = 1e{-2}\) and \(\lambda = 1e{-5}\) provide the best trade-off between watermark detectability and fluency. In contrast, aggressive fine-tuning with \(\lambda = 0.1\) leads to excessive watermark embedding, resulting in lower precision and noticeable fluency degradation. Exponential regularized barriers outperform quadratic ones by ensuring a more stable optimization process, as reflected in their higher ROC-AUC and PR-AUC scores.See Appendix \ref{appendix:results} for additional results on Accuracy,F1-score and Precision across datasets.


These results establish a robust framework for watermarking open-weight language models and highlight the importance of regularized optimization in balancing watermark robustness with linguistic naturalness.









\section{Conclusion}

We introduced an end-to-end watermarking framework by jointly fine-tuning LoRA adapters within a LLaMA 3.1 8B model, utilizing a performer model for text generation and an observer model for watermark detection. By optimizing the Binoculars framework, we ensured that watermark embedding remains both effective and subtle while preserving text coherence.

Our results demonstrate that structured regularization prevents over-optimization, enabling a controlled watermarking process that maintains fluency. Models trained with \(\lambda = 1e{-2}\) and \(1e{-5}\) achieved optimal performance, balancing detectability and naturalness without degrading accuracy. The findings highlight the advantage of training-integrated watermarking, which enhances robustness without relying on inference-time modifications.

Future work should explore how this framework generalizes across different LLM architectures and its resistance to text transformations such as paraphrasing and adversarial perturbations. Further research is also needed to refine watermark verification in real-world applications, ensuring practical deployment in AI accountability and content attribution.




\bibliographystyle{plainnat} 

\newpage
\appendix
\onecolumn
\section{Further Trials and Experiments}

\subsection{Alternative Loss Formulations}

During training, we explored different loss formulations to balance watermarking detectability and text fluency. One key variation involved modifying the role of the binocular score for generated sequences. By adjusting its contribution, we aimed to encourage more controlled watermark embedding:

\begin{equation}
    L_{\text{total}} = L_{\mathcal{M}_P}(s) + \lambda \cdot \mathcal{B}_{\mathcal{M}_O, \mathcal{M}_P}(s) - \mathcal{B}_{\mathcal{M}_O, \mathcal{M}_P}(s_{\text{gen}}).
\end{equation}

This formulation treats the binocular score of the generated sequence as an opposing term, refining how watermarking is applied during training.

Further refinements incorporated cross-perplexity (\(\operatorname{XPPL}\)), which measures the alignment between the observer’s and performer’s predictions. By including this metric, we sought to maintain coherence while optimizing watermark embedding:

\begin{equation}
    L_{\text{total}} = L_{\mathcal{M}_P}(s) + \operatorname{XPPL}_{\mathcal{M}_O, \mathcal{M}_P}(s) - \operatorname{XPPL}_{\mathcal{M}_O, \mathcal{M}_P}(s_{\text{gen}}).
\end{equation}

This adjustment ensures that the observer’s probability distribution remains calibrated with the performer’s, promoting subtle yet effective watermarking.

Finally, we tested an alternative formulation where the sign of the cross-perplexity term was flipped:

\begin{equation}
    L_{\text{total}} = L_{\mathcal{M}_P}(s) - \lambda \cdot \operatorname{XPPL}_{\mathcal{M}_O, \mathcal{M}_P}(s) + \operatorname{XPPL}_{\mathcal{M}_O, \mathcal{M}_P}(s_{\text{gen}}).
\end{equation}

This approach modifies the optimization dynamics, shifting how the model aligns predictions for watermark embedding while ensuring the generated text remains coherent.

These experiments provided insights into the impact of different training objectives, highlighting trade-offs between watermark robustness and naturalness. While not used in the final implementation, they offer alternative directions for future work in optimizing watermarking strategies.

\section{Alternative Optimization Strategies}\label{appendix:optimization}
We explore different optimization strategies to balance watermark embedding and detectability.

\paragraph{Hybrid Loss on Sequences}
This approach optimizes both the performer and observer using:
\begin{equation}
    \min_{\mathcal{M}_P, \mathcal{M}_O} L_{\text{total}} = L_{\mathcal{M}_P}(s) -  \lambda \cdot \mathcal{B}(s) + \mathcal{B}(s_{\text{gen}}).
\end{equation}
Here, the observer evaluates both human and generated sequences to maintain watermark detectability across distributions.

\paragraph{Separate Optimization}
An alternative decouples the objectives:
\begin{equation}
\begin{aligned}
    \min_{\mathcal{M}_P} &\left(L_{\mathcal{M}_P}(s) -  \lambda \cdot \mathcal{B}(s)\right), \\
    \min_{\mathcal{M}_O} &\mathcal{B}(s_{\text{gen}}).
\end{aligned}
\end{equation}
This lets the performer focus on embedding watermarks while the observer refines detection.

\paragraph{Cross-Perplexity Alignment}
We also leverage cross-perplexity (\(\operatorname{XPPL}\)) to align the observer’s predictions with generated text:
\begin{equation}
\begin{aligned}
    \min_{\mathcal{M}_P} &\left(L_{\mathcal{M}_P}(s) + \operatorname{XPPL}(s)\right), \\
    \min_{\mathcal{M}_O} &-\operatorname{XPPL}(s_{\text{gen}}).
\end{aligned}
\end{equation}

These variants highlight different trade-offs in optimizing detectability without degrading text quality.


\section{Additional Results}
\label{appendix:results}
\begin{figure}[t]
    \centering
    \begin{minipage}{0.32\textwidth}
        \centering
        \includegraphics[width=\linewidth]{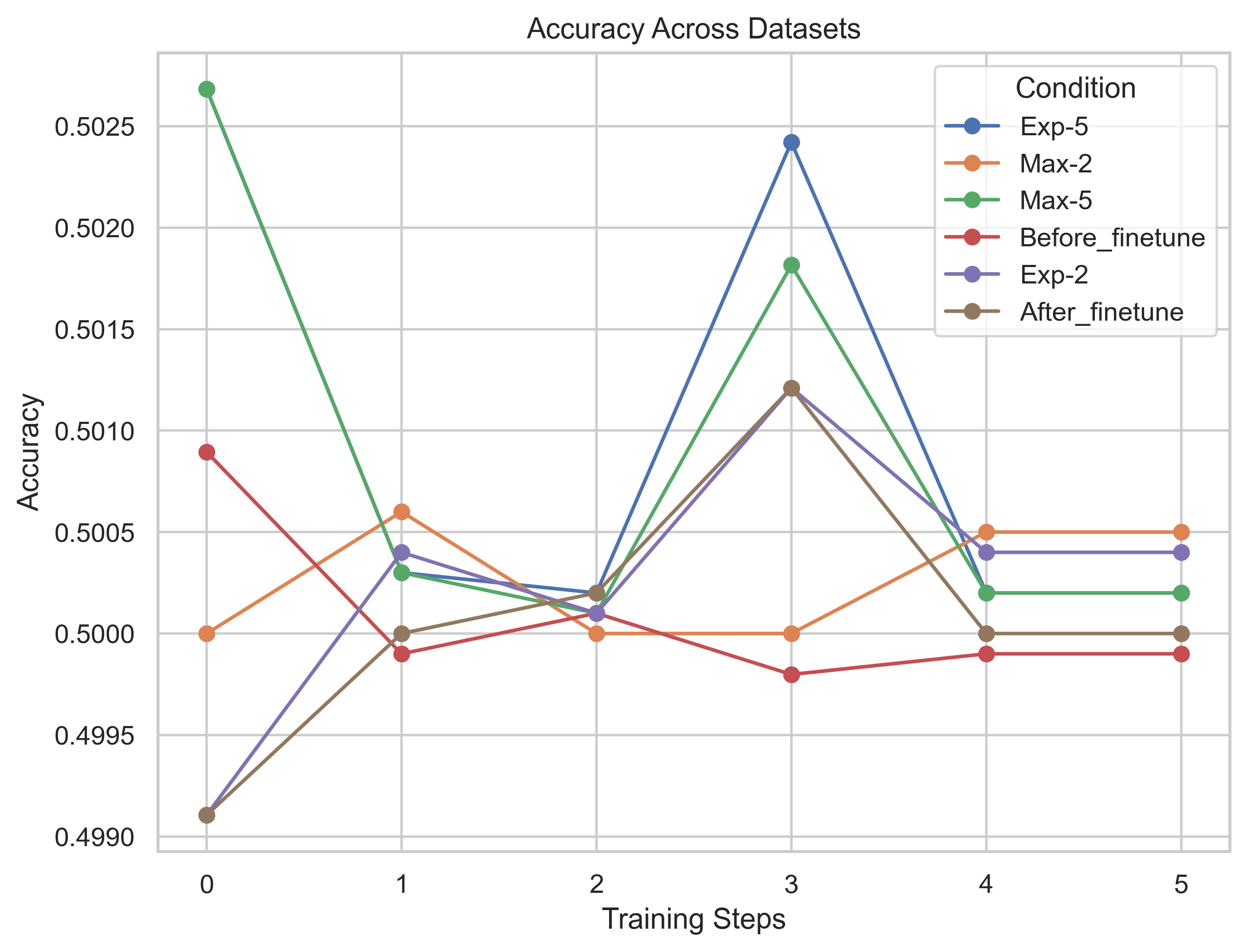}
        \subcaption{Accuracy trends. Exponential regularization (\(\lambda = \num{1e-5}, \num{1e-2}\)) stabilizes training.}
        \label{fig:accuracy}
    \end{minipage}
    \hfill
    \begin{minipage}{0.32\textwidth}
        \centering
        \includegraphics[width=\linewidth]{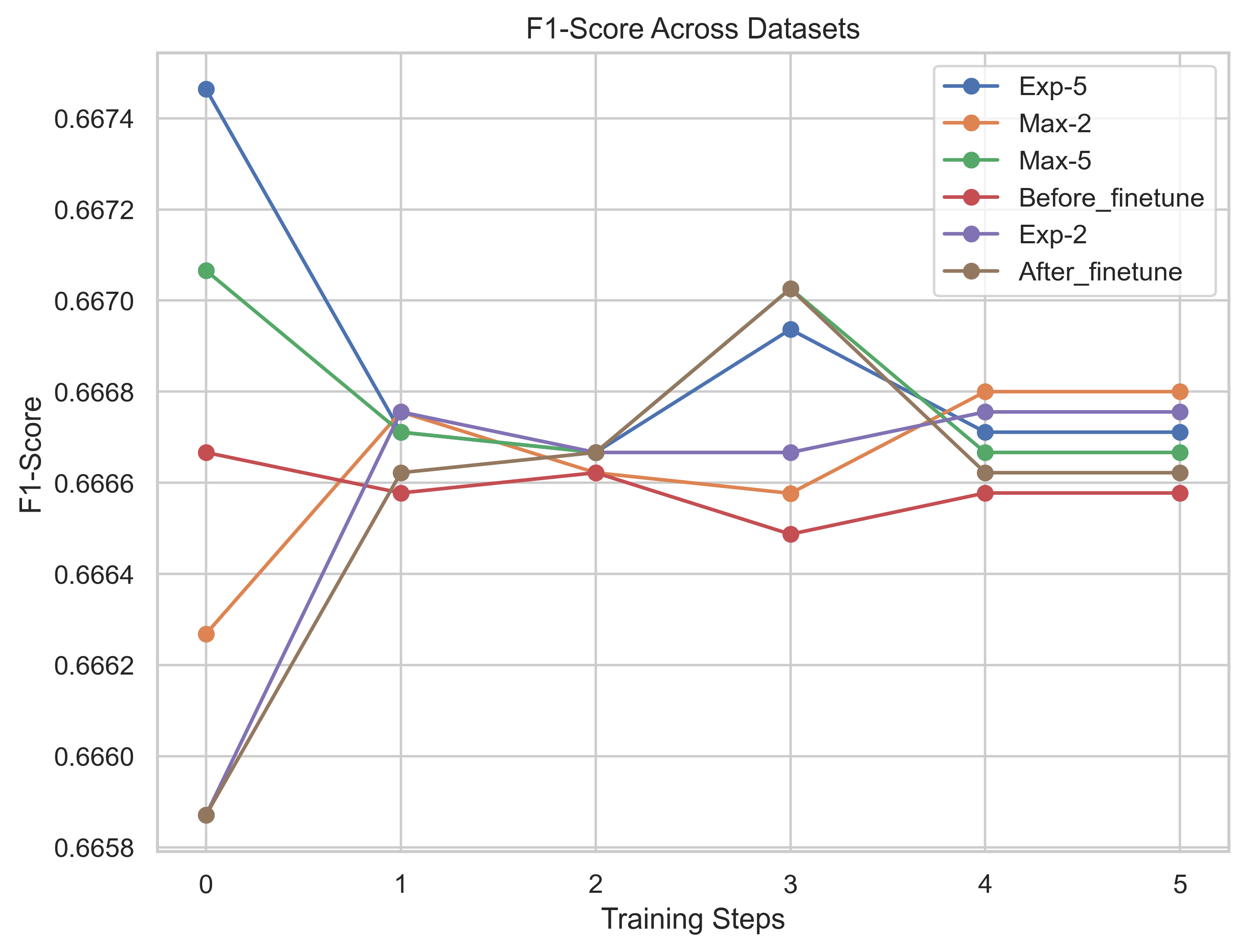}
        \subcaption{F1-score trends. Regularized models maintain higher consistency.}
        \label{fig:f1_score}
    \end{minipage}
    \hfill
    \begin{minipage}{0.32\textwidth}
        \centering
        \includegraphics[width=\linewidth]{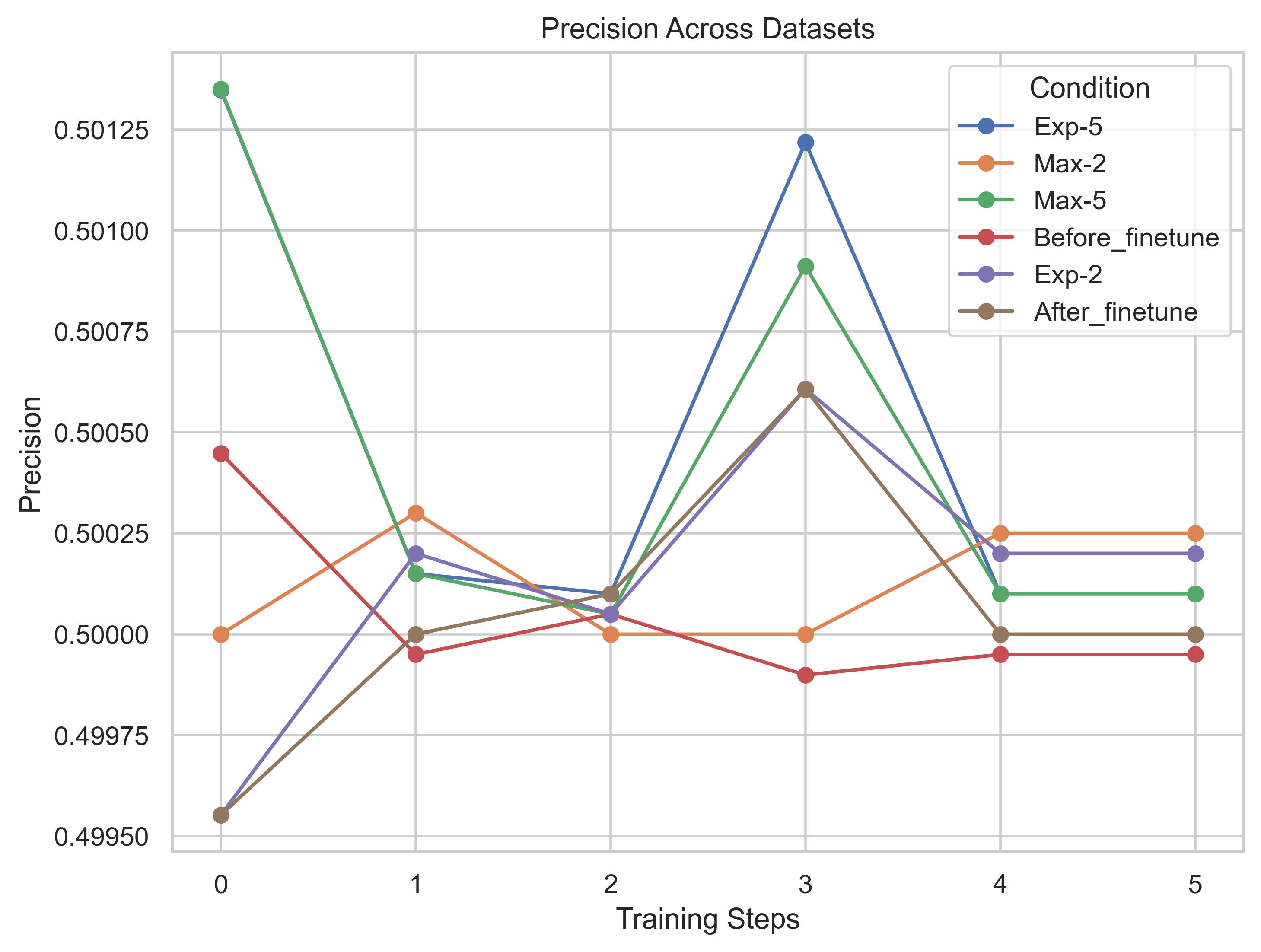}
        \subcaption{Precision trends. Models with \(\lambda = \num{1e-5}, \num{1e-2}\) exhibit higher precision.}
        \label{fig:precision}
    \end{minipage}
    \caption{Comparison of regularized models on accuracy, F1-score, and precision. Exponential regularization improves detectability while ensuring training stability.}
    \label{fig:performance_comparison}
\end{figure}

Table \ref{tab:appendix_thresholds_study} extends the evaluation of threshold values and F1 scores across Wiki and CC News datasets, reinforcing the impact of different regularization strategies. The exponential regularized barrier with \(\lambda = 1e^{-5}\) achieves the highest FPR threshold (4.1480 for Wiki), enhancing watermark detectability. The maximum-based quadratic barrier (Max\_1e-5) attains the highest accuracy thresholds, preserving performance while embedding watermarks. The baseline model shows the lowest accuracy, confirming that standard LoRA fine-tuning alone is insufficient for strong watermark separation. These results further validate the effectiveness of controlled regularization. See Figure \ref{fig:performance_comparison}

\begin{table}[!htbp]
    \centering
    \footnotesize
    \caption{Threshold values and F1 scores for Wiki and CC News datasets. 
    The table compares different experimental conditions based on accuracy threshold, false positive rate (FPR) threshold, and best F1 score.}
    \label{tab:appendix_thresholds_study}
    \vskip 0.15in
    \begin{tabular}{l c c c c c c c c}
        \toprule
        \multirow{2}{*}{\textbf{Condition}}
        & \multicolumn{4}{c}{\textbf{Wiki Dataset}} 
        & \multicolumn{4}{c}{\textbf{CC News Dataset}} \\
        \cmidrule(lr){2-5} \cmidrule(lr){6-9}
        & \textbf{Accuracy} & \textbf{FPR} & \textbf{Best F1} & \textbf{FPR-TPR} 
        & \textbf{Accuracy} & \textbf{FPR} & \textbf{Best F1} & \textbf{FPR-TPR} \\
        \midrule
        \texttt{Gradexp\_lam0.1} 
        & 1.2128 & 0.8953 & \textbf{0.6669} & 0.0000  
        & 0.8430 & 0.6155 & \textbf{0.6668} & 0.0000 \\
        \texttt{explam\num{1e2}}      
        & 0.9414 & 0.7442 & 0.6668 & 0.0000  
        & 0.5895 & 0.4771 & \textbf{0.6668} & 0.0000 \\
        \texttt{maxlam\num{1e5}}      
        & \textbf{1.3057} & 4.0506 & 0.6668 & 0.0004  
        & \textbf{1.1307} & \textbf{3.1896} & \textbf{0.6668} & 0.0008 \\
        \texttt{explam\num{1e5}}      
        & 1.1409 & \textbf{4.1480} & 0.6668 & \textbf{0.0008}  
        & 0.9829 & 3.1023 & \textbf{0.6668} & \textbf{0.0010} \\
        \texttt{basebefore}      
        & 0.6952 & 0.6952 & 0.6667 & 0.0000  
        & 0.5590 & 1.5557 & 0.6667 & 0.0002 \\
        \bottomrule
    \end{tabular}
    \vskip -0.1in
\end{table}

\end{document}